\documentclass[aps,pre,twocolumn,superscriptaddress,floatfix]{revtex4-2}

\usepackage{graphicx}
\usepackage{xcolor}
\usepackage{amsmath}
\usepackage{amssymb}
\usepackage{amsfonts}
\usepackage{bm}
\usepackage{booktabs}
\usepackage{hyperref}
\hypersetup{
    colorlinks=true,
    linkcolor=blue,
    citecolor=blue,
    urlcolor=blue,
}

\newcommand{\smax}{s_{\mathrm{max}}}
\newcommand{\savg}{\langle s \rangle}
\newcommand{\Nparams}{N}

\begin{document}

\title{Dimensional Criticality at Grokking Across MLPs and Transformers}

% NOTE: Replace with the actual author list before submission.
\author{Ping Wang}
\affiliation{Institute of High Energy Physics, Chinese Academy of Science, 
100049 Beijing, China}
\email{pwang@ihep.ac.cn}

\date{\today}

\begin{abstract}
Abrupt transitions between distinct dynamical regimes are a hallmark of
complex systems.
Grokking in deep neural networks provides a striking example---an
abrupt transition from memorization to generalization long after
training accuracy saturates---yet robust macroscopic signatures of
this transition remain elusive.
Here we introduce \textbf{TDU--OFC} (Thresholded Diffusion Update--Olami-Feder-Christensen),
an offline avalanche probe that converts gradient snapshots
into cascade statistics and extracts a \emph{macroscopic observable}---the
time-resolved effective cascade dimension $D(t)$---via grokking-aligned
finite-size scaling.
Across Transformers trained on modular addition and MLPs trained on XOR, we
discover a localized dynamical crossing of the Gaussian diffusion baseline $D=1$
precisely at the generalization transition.
The crossing direction is task-dependent: modular addition descends through
$D=1$ (approaching from $D>1$), while XOR ascends (from $D<1$). This
opposite-direction convergence is consistent with attraction toward a candidate
shared critical manifold, rather than trivial residence near $D \approx 1$.
Negative controls confirm this picture: ungrokked runs remain supercritical
($D>1$) and never enter the post-transition regime.  In addition, avalanche 
distributions exhibit heavy tails and finite-size scaling consistent with 
the dimensional exponent extracted from $D(t)$.
Shadow-probe controls ($\alpha_{\mathrm{train}}=0$) confirm that $D(t)$ is
non-invasive, and grokked trajectories diverge from ungrokked ones in $D(t)$
some $100$--$200$ epochs before the behavioral transition.

\end{abstract}

\maketitle

\section{Introduction}

Abrupt transitions between qualitatively distinct dynamical regimes are a
hallmark of complex systems~\cite{goldenfeld1992lectures,bak1987self}.
Recent work in deep learning has revealed a striking instance of this
phenomenon---grokking~\cite{power2022grokking}---in which neural networks
abruptly transition from memorization to generalization: training accuracy
saturates early while test accuracy remains poor, followed by a delayed
and abrupt generalization breakthrough.
This behavior has been interpreted as a sharp change in effective learning
dynamics akin to a phase transition in statistical physics.
In well-studied critical phenomena, such abrupt reorganizations are
accompanied by \emph{macroscopic observables}---order parameters---that
signal the transition and exhibit scale invariance near the critical
point, without reference to microscopic degrees of freedom~\cite{goldenfeld1992lectures}.
An analogous macroscopic order parameter for grokking, however, remains
elusive despite growing mechanistic understanding.

A central challenge is identifying macroscopic observables capable of
predicting \emph{when} this transition occurs and characterizing the
collective reorganization of gradient geometry that drives it.
Circuit-level mechanistic interpretability~\cite{nanda2023progress} reveals
\emph{which} internal features emerge during grokking; complementary
theoretical analyses propose effective descriptions of representation
learning~\cite{liu2022grokking}.
Yet these approaches do not provide macroscopic observables capable of
forecasting the transition---tracking the collective reorganization of
optimization geometry during learning requires a complementary physics-based
approach.
A physics-based perspective can address a distinct question: \emph{how} does
the collective geometry of optimization reorganize during generalization?
The two approaches probe the same transition at different scales of description.
The present work targets a different layer of description: a statistical-physics
characterization of collective geometric reorganization at the generalization
transition, complementary to microscopic interpretability approaches.

The physics-based perspective suggests that training dynamics may 
self-organize into a critical state characterized by scale invariance 
and heavy-tailed avalanches, a hallmark of self-organized criticality 
(SOC)~\cite{bak1987self,pruessner2012self}.
SOC arises in diverse driven dissipative systems---including earthquakes 
modeled by thresholded redistribution dynamics~\cite{olami1992self}---and 
yields power-law event-size distributions and finite-size scaling (FSS) 
relations.
Avalanche dynamics also underlie crackling noise in many physical 
systems~\cite{sethna2001crackling}, and have been discussed in biological 
contexts such as neuronal avalanches and brain criticality~\cite{beggs2003neuronal,chialvo2010emergent,munoz2018criticality}.
If neural network training exhibits SOC-like cascades in gradient space, 
then grokking may correspond to a transient critical regime in which the 
effective exploration geometry reorganizes.
Relatedly, analytic results in simplified regimes (deep linear networks 
and the neural tangent kernel limit) provide tractable baselines for 
learning dynamics~\cite{saxe2014exact,jacot2018neural}, while empirical 
studies highlight rich structure in neural loss landscapes~\cite{li2018visualizing,fort2019deep} 
and heavy-tailed self-regularization signatures~\cite{martin2019traditional}.
Recent work has also identified quasi-critical avalanche dynamics with 
distinct universality classes in deep networks during 
training, identifying quasi-critical activation-cascade 
dynamics~\cite{ghavasieh2025quasi} and tunable information-propagation 
universality classes~\cite{ghavasieh2025tunable} during learning.
Our approach is complementary: rather than proposing a new solvable limit,
we operationalize SOC diagnostics specifically in gradient space, aligned to
the grokking transition event.
Unlike activation- and signal-propagation probes~\cite{ghavasieh2025quasi,ghavasieh2025tunable},
$D(t)$ directly captures the gradient correlation geometry at the
generalization transition, and we provide a reproducible measurement
protocol---grokking-aligned, time-resolved FSS---that localizes the phase
transition in time and admits negative controls.

To operationalize SOC diagnostics in gradient space, we require a diagnostic 
that converts gradient snapshots into a quantifiable correlation signature.
During backpropagation, gradients across different parameters acquire 
correlations through shared loss landscape structure and the chain 
rule~\cite{jacot2018neural}.
When these correlations are strong, a perturbation in one gradient 
component propagates to many others---analogous to how the correlation 
length diverges at a phase transition in spin 
systems~\cite{goldenfeld1992lectures}.
TDU--OFC (Thresholded Diffusion Update--Olami-Feder-Christensen) operationalizes
this intuition: real training gradients are injected as initial conditions
into a threshold-driven OFC-style diffusion process, and we measure how far perturbations cascade across a 
parameter-space graph, analogous to extracting effective spatial dimension 
from tracer diffusion in fractal media.
The mean avalanche size acts as a \emph{generalized 
susceptibility}~\cite{pruessner2012self,sethna2001crackling} whose growth 
with system size signals criticality; $D(t)$, extracted via FSS, tracks the geometry of this susceptibility over training 
time and provides a physically interpretable macroscopic readout of 
gradient correlation structure.
$D(t)$ thus serves as a \emph{macroscopic order parameter} for the
gradient-geometry phase transition at grokking.

In a recent study~\cite{wang2025grokking_xor}, we established that XOR
grokking under gradient descent exhibits SOC signatures with
quasi-one-dimensional exponents $D\approx 1.06$.
Here we show that \emph{grokking corresponds to a dimensional critical
transition in gradient cascade geometry}: $D(t)$, extracted offline from
gradient snapshots, undergoes a characteristic \emph{dynamical crossing}
through the Gaussian baseline $D=1$ precisely at the generalization
transition, and diverges between grokked and ungrokked trajectories
$100$--$200$ epochs before behavioral grokking.
We establish this across (i) Transformers trained on
modular arithmetic (ModAdd), a canonical grokking benchmark~\cite{power2022grokking},
and (ii) MLPs trained on XOR, which provides a contrasting architecture and
task family.
The present study makes three contributions:
(i) \emph{shadow-probe validation} ($\alpha_{\mathrm{train}}=0$,
$\max|\Delta D|\leq 0.14$) confirming that $D(t)$ remains stable under
$\sim30^\circ$ local gradient rotations introduced by the probe,
establishing it as a robust macroscopic observable of gradient correlation geometry;
(ii) \emph{grokking-aligned, time-resolved FSS} as a general protocol for 
detecting critical transitions in asynchronous learning systems where 
grokking epochs span orders of magnitude across scales; and
(iii) the \emph{directional $D=1$ crossing} at grokking: ModAdd descends through
$D=1$ while XOR ascends, with shadow-probe controls confirming the signature
is intrinsic and opposite crossing directions providing evidence against task-specific artifacts.
\emph{Together, these results provide evidence that grokking corresponds to a
dimensional critical transition in gradient cascade geometry: $D(t)$ crosses
the Gaussian baseline $D=1$ at the generalization boundary, the crossing
direction encodes task structure, and the critical manifold $D=1$ is consistent
with a candidate shared attractor across the two architectures and task families studied.}
We discuss this as a candidate \emph{Dimensional Grokking Principle} in the Discussion.

\section{Methods}

\paragraph{Tasks and models.}
We study modular addition (ModAdd-$p$) with $p=59$: the network receives 
$(a,b)$ and predicts $(a+b)\bmod p$ as a $p$-way classification.
The $p^2=3{,}481$ input pairs are split 80/20 into train/test sets, producing 
the canonical delayed generalization (``grokking'') behavior~\cite{power2022grokking}.

For ModAdd we use Transformer encoders~\cite{vaswani2017attention}
with token and positional embeddings and a last-token classification head.
We sweep the system size $\Nparams$ by varying $d_{\mathrm{model}}$, while
keeping the number of attention heads fixed at $n_{\mathrm{heads}}=4$ for FSS
unless otherwise stated. For ModAdd we use Adam~\cite{kingma2015adam}
(lr $5\times10^{-3}$, weight decay $10^{-3}$).
Grokking time $g$ is defined as the first epoch at which the test accuracy
exceeds $0.99$ (measured at snapshot intervals).

For comparison, we also analyze the XOR task with a two-layer MLP 
(varying hidden width to sweep $\Nparams$). 
XOR has only 4 input patterns with no train/test split; 
grokking is identified when accuracy on all samples exceeds 99\%.
Since training and evaluation use the same four patterns, the transition
is an abrupt learning event in gradient geometry rather than canonical
delayed generalization~\cite{wang2025grokking_xor}---a methodological
feature that isolates the gradient-level phase transition from
train/test behavioral confounds.

\paragraph{TDU--OFC avalanche probe.}
We apply a thresholded diffusion update inspired by the OFC earthquake 
model~\cite{olami1992self,ChristensenOlami1992OFC}.
For each batch, we collect the gradient $\bm{g}\in\mathbb{R}^{\Nparams}$, define 
threshold $\tau$ as the 90th percentile of $|\bm{g}|$, and construct a fixed 
Barab\'asi--Albert graph ($m=2$)~\cite{barabasi1999emergence} over parameters.
The BA graph provides an architecture-agnostic substrate: its
scale-free degree distribution captures the heterogeneous
influence of parameters on gradient magnitudes without imposing
any specific network topology; control experiments below confirm
that $D$ is insensitive to this choice.
Nodes with $|g_i|>\tau$ are ``active'' and trigger OFC-style redistribution: 
the active node $i$ decays and redistributes equally to its $k_i$ neighbors,
\begin{equation}
g_i' = (1-\alpha)\,g_i, \quad 
g_j' = g_j + \frac{\alpha\,g_i}{k_i} \;\;\text{for all } j\sim i,
\label{eq:diffusion}
\end{equation}
with $\alpha=0.3$ for up to 20 cascade iterations.
The avalanche size $s$ is the total number of active update events across all iterations.
The degree normalization in Eq.~\eqref{eq:diffusion} ensures quasi-conservative 
redistribution---analogous to the quasi-conservative limit of the OFC 
model~\cite{olami1992self}---so that cascade extent reflects gradient 
correlation structure rather than unbounded energy injection.
This implements the condensed-matter paradigm of probing internal 
correlations through relaxation response: the avalanche is the system's 
relaxation to above-threshold gradient perturbations, and its size 
quantifies the spatial extent of this relaxation in parameter space.
The mean avalanche size $\langle s \rangle$ acts as a \emph{generalized 
susceptibility}~\cite{pruessner2012self} whose growth with $\Nparams$ 
signals criticality; we use $\smax$ as the primary FSS observable due 
to its sensitivity to the largest correlated cascade, with $\langle s \rangle$ 
providing consistent supporting behavior.

Control experiments with synthetic i.i.d.\ Gaussian 
gradients (Fig.~\ref{fig:modadd_bootstrap_loo}a; also~\cite{wang2025grokking_xor})
yield $D_{\mathrm{synth}}\approx 1$ invariant to graph topology
(coefficient of variation $<0.4\%$), confirming 
that $D$ reflects backpropagation correlation geometry rather than 
the diffusion graph structure.
The threshold $\tau$ at the 90th percentile selects the top 10\% most
active gradient components, balancing sensitivity to large-magnitude
updates against noise suppression; the redistribution fraction $\alpha=0.3$ is a moderate value that
allows cascade propagation while preserving local gradient structure;
since the degree normalization in Eq.~\eqref{eq:diffusion} ensures
exact conservation for any $\alpha$ on the BA graph, this parameter
controls cascade intensity rather than energy conservation, unlike
the classical 2D OFC model~\cite{olami1992self}.
For XOR, a systematic parameter sweep ($\alpha\in\{0.2,0.3,0.4,0.5\}$,
$\tau\in\{85\mathrm{th},90\mathrm{th},95\mathrm{th}\}$ percentile,
3~seeds per configuration, 36~runs total) confirms robustness:
grokking timing varies by ${<}3\%$ across the tested parameter space
($p>0.7$, not significant), with threshold percentile showing
negligible effect across the entire range tested.
These results indicate that SOC emergence is governed primarily
by intrinsic gradient statistics rather than diffusion mechanics,
consistent with the self-organizing nature of critical phenomena.

\paragraph{Shadow-probe control.}
To verify that the probe does not create SOC signatures, we run a shadow-probe 
control ($\alpha_{\mathrm{train}}=0$): parameters update via raw gradients 
while avalanches are computed offline from copied gradients, ensuring 
that the avalanche diagnostic does not influence the training dynamics.

\paragraph{Temporal aggregation and FSS.}
We coarse-grain batch avalanches into epoch statistics: for each epoch we 
compute $\smax$ (maximum per-batch avalanche size within the epoch) and $\savg$ (mean per-batch avalanche size), then fit 
$\smax \sim \Nparams^{D}$ and $\savg \sim \Nparams^{\gamma}$ via log--log 
regression.
We focus on $D(t)$ from $\smax$ as the primary diagnostic; $\gamma(t)$ shows 
consistent behavior~\cite{wang2025grokking_xor}.
Unless otherwise noted, the main FSS analysis uses grokked runs; 
ungrokked runs are analyzed separately as negative controls.
To compare dynamics across scales, we use grokking-aligned relative time 
$t=(\mathrm{epoch}-g)/g$.

For distributional evidence we define epoch-summed avalanches 
$S_{\mathrm{epoch}}=\sum_b s_b$ and analyze their cumulative distribution 
within $\pm 500$ epochs of grokking.
For cutoff scaling (Fig.~\ref{fig:cdf}), we pool $S_{\mathrm{epoch}}$ across 
epochs and seeds, define $s_c$ as the 95th percentile, and fit 
$s_c\sim N^{D_{\mathrm{cut}}}$.

\paragraph{Grokking-aligned FSS.}
Grokking epochs span two orders of magnitude ($g \in [29, 3000]$ for
$d_{\mathrm{model}} \in [16, 128]$); the aligned coordinate $t$ introduced
above is therefore essential for separating pre- and post-grokking phases
across scales, and provides a universally applicable window for any abrupt
generalization phenomenon.

\section{Results}

The following results are organized as a five-layer evidence chain
supporting the interpretation of grokking as a dimensional critical transition
in gradient cascade geometry.
\textbf{R1} (Sec.~\ref{sec:r1}) establishes the canonical behavioral signature
of grokking and motivates the grokking-aligned time coordinate $t=(\mathrm{epoch}-g)/g$.
\textbf{R2--R3} (Sec.~\ref{sec:r2}) build the core case for dimensional criticality:
\textbf{R2} demonstrates the directional $D=1$ crossing in grokking-aligned relative time
(Fig.~\ref{fig:temporal_fss}); \textbf{R3} provides independent distributional
evidence via heavy-tailed finite-size scaling (Fig.~\ref{fig:cdf}) and
bootstrap resampling and leave-one-out analysis quantifying the phase separation
($D_{\mathrm{pre}}=1.12\pm0.02$, $D_{\mathrm{post}}=0.92\pm0.02$;
Fig.~\ref{fig:modadd_bootstrap_loo}).
\textbf{R4} (Sec.~\ref{sec:r3}) provides negative controls via ungrokked
trajectories, which remain persistently supercritical, and demonstrates
robustness of the $D=1$ crossing across modular primes.
\textbf{R5} (Sec.~\ref{sec:r4}) completes the diagnostic protocol with
shadow-probe non-invasiveness controls.
Together, these five layers constitute a self-consistent, falsifiable characterization
of dimensional criticality at the generalization transition.

\subsection{Canonical grokking delay and grokking-aligned time}
\label{sec:r1}
\begin{figure}[tp]
\centering
\includegraphics[width=\columnwidth]{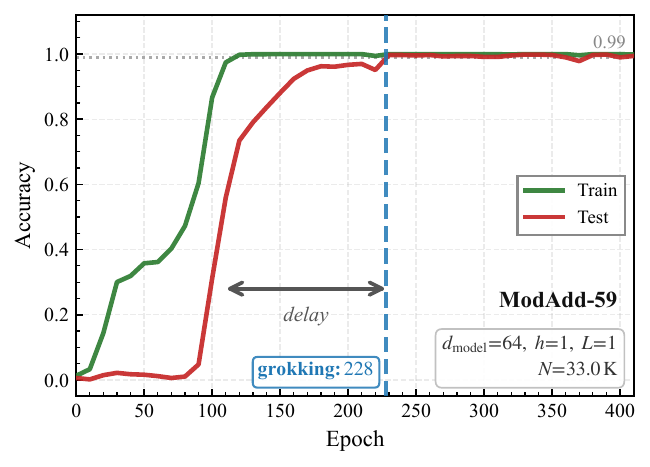}
\caption{\textbf{Canonical grokking delay and definition of the transition epoch $g$ in ModAdd-59.}
A representative run shows early training saturation and delayed test
generalization. The dashed vertical line marks the grokking epoch $g$ and the
double-headed arrow indicates the train-test delay interval.}
\label{fig:grok_delay}
\end{figure}

We first establish the canonical behavioral signature of grokking: an extended
train-test delay in which training accuracy saturates early while test accuracy
remains low until a late generalization breakthrough. Figure~\ref{fig:grok_delay}
shows a representative ModAdd-59 trajectory.

Because grokking times span orders of magnitude across system sizes, analyses on
an absolute epoch axis can mix pre- and post-transition regimes across scales.
To isolate transition-localized signatures, we align dynamics by the grokking
epoch $g$ and use the relative coordinate $t=(\mathrm{epoch}-g)/g$
(Methods).

\subsection{Dimensional criticality at grokking}
\label{sec:r2}
\begin{figure}[tp]
\centering
\includegraphics[width=\columnwidth]{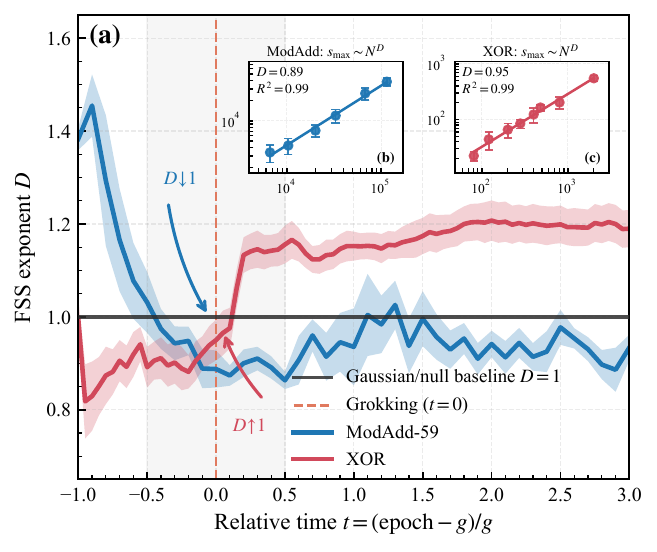}
\caption{\textbf{Dynamical dimensional criticality at grokking.}
\textbf{(a)} Time-resolved FSS exponent $D(t)$ extracted from
$\smax\sim\Nparams^{D}$, plotted against normalized time
$t=(\mathrm{epoch}-g)/g$ aligned to each run's grokking epoch $g$.
ModAdd-59 descends through $D=1$ near grokking, whereas XOR ascends, with both
crossings localized at $t\approx 0$.
The shaded band marks the critical window $|t|\leq 0.5$.
\textbf{(b,c)} Log--log fits at $t\approx 0$ for ModAdd-59 and XOR confirm
consistent finite-size scaling quality in the transition window.}
\label{fig:temporal_fss}
\end{figure}

Having aligned all scales to their transitions, we extract the time-resolved FSS
exponent $D(t)$ from $\smax\sim \Nparams^{D}$. Figure~\ref{fig:temporal_fss}
shows a robust, grokking-localized \emph{dynamical crossing} through the
Gaussian/null baseline $D=1$: ModAdd-59 descends through $D=1$ while XOR
ascends. Because uncorrelated Gaussian gradients on the same topology can also
yield $D\approx 1$~\cite{wang2025grokking_xor}, our operational criterion is
the transition-localized \emph{crossing} rather than static residence near
$D\approx 1$.

\begin{figure}[tp]
\centering
\includegraphics[width=\columnwidth]{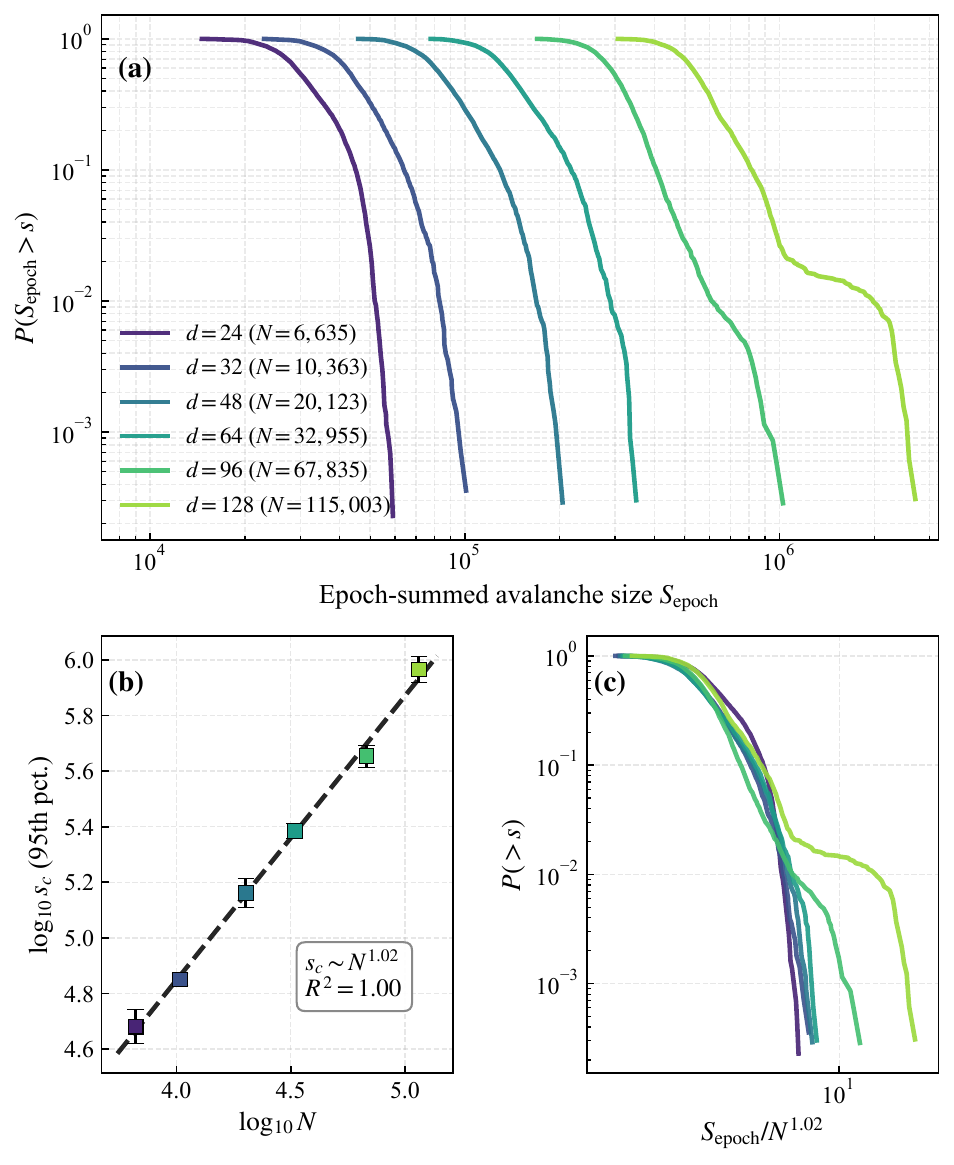}
\caption{\textbf{Heavy-tailed avalanches with finite-size scaling in ModAdd-59.}
\textbf{(a)} CCDF of epoch-summed avalanches $S_{\mathrm{epoch}}$ (within $\pm 500$
epochs of grokking) for six model widths.
\textbf{(b)} Cutoff scaling $s_c \sim N^{D_{\mathrm{cut}}}$ with
$D_{\mathrm{cut}} \simeq 1.02$ ($R^2 \simeq 1.00$), where $s_c$ is the
95th-percentile cutoff of the pooled $S_{\mathrm{epoch}}$ distribution.
\textbf{(c)} Data collapse: rescaling by $S_{\mathrm{epoch}}/N^{D_{\mathrm{cut}}}$
collapses the distributions.}
\label{fig:cdf}
\end{figure}

\begin{figure}[tp]
\centering
\includegraphics[width=\columnwidth]{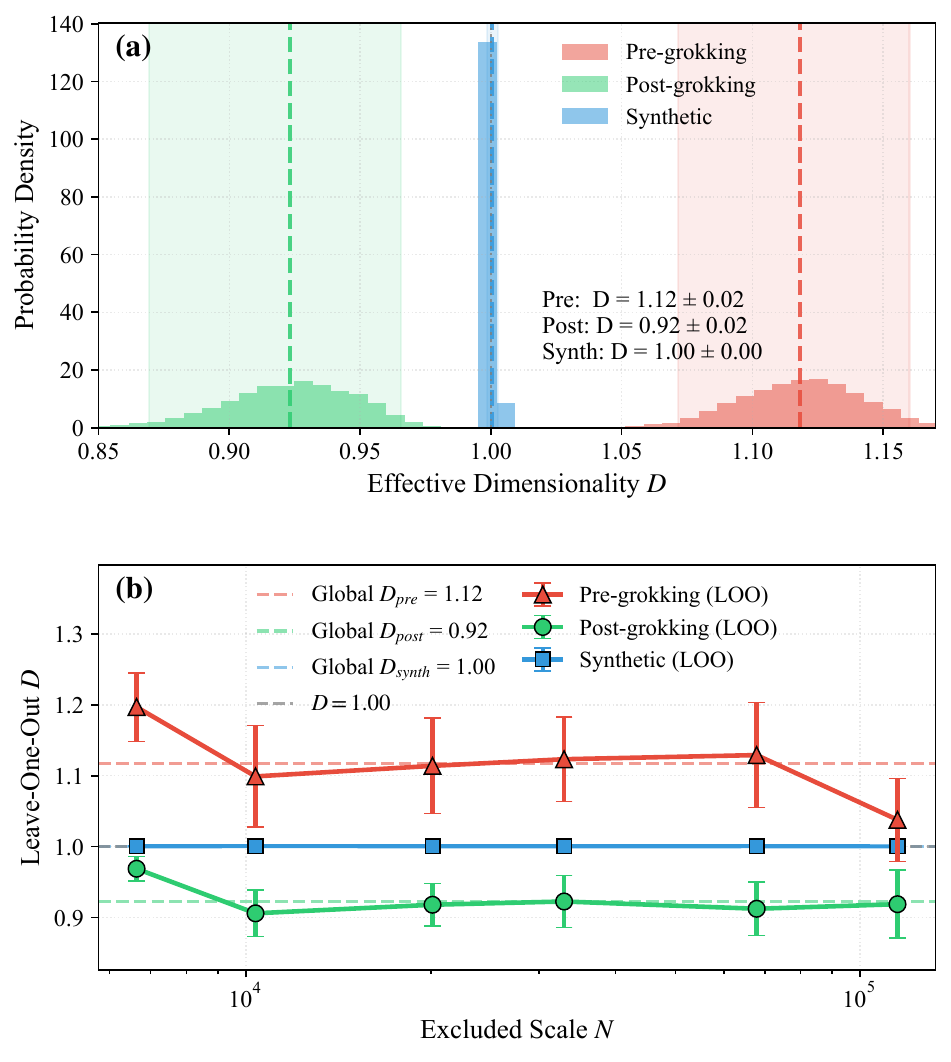}
\caption{\textbf{Statistical robustness of the pre/post-grokking $D$ separation in ModAdd-59.}
\textbf{(a)} Bootstrap distributions of the phase-split dimensional exponent $D$
(5000 resamples, strict $t=0$ split).
Pre-grokking ($D_{\mathrm{pre}}=1.12\pm 0.02$) and post-grokking 
($D_{\mathrm{post}}=0.92\pm 0.02$) are well-separated from each other and 
from the scale-free Gaussian synthetic baseline ($D_{\mathrm{synth}}=1.00\pm 0.00$).
\textbf{(b)} Leave-one-scale-out (LOO) analysis: removing any single intermediate
scale ($N=10{,}363$--$67{,}835$) shifts $D$ by ${<}0.02$; excluding either extreme
scale introduces a shift of at most $0.08$, consistent with their larger leverage
in the power-law regression.  Both phases remain well-separated from $D=1$ under
all six leave-one-out variants, confirming the robustness of the phase separation.}
\label{fig:modadd_bootstrap_loo}
\end{figure}

Complementary \emph{distributional criticality}---heavy-tailed, scale-free
avalanche statistics with finite-size scaling (FSS)---provides independent
evidence for SOC-like regimes~\cite{pruessner2012self}.
Figure~\ref{fig:cdf} analyzes epoch-summed avalanche sizes
$S_{\mathrm{epoch}}=\sum_{b\in\mathrm{epoch}} s_b$ within a $\pm 500$ epoch
window around each run's grokking time (Methods); this window focuses on
grokking-critical dynamics and excludes post-convergence fluctuations at late
epochs that can dominate the tail statistics and distort the FSS relationship.
The finite-size cutoff
$s_c$ (95th percentile) follows a clean power law $s_c \sim N^{D_{\mathrm{cut}}}$
with $D_{\mathrm{cut}} \simeq 1.02$ and $R^2 \simeq 1.00$, and the distributions
collapse under rescaling.
The value $D_{\mathrm{cut}}\simeq 1.02$ is intermediate between the
pre- and post-grokking phase estimates (to be quantified in the bootstrap analysis
below): because the $\pm 500$ epoch analysis window spans both sides of
the grokking transition, $D_{\mathrm{cut}}$ is a \emph{time-averaged} exponent
over two distinct scaling regimes~\cite{wang2025grokking_xor}, not a single
stationary exponent.
The residual spread in the data collapse (Fig.~\ref{fig:cdf}c) is a direct
signature of this non-stationarity, motivating the phase-resolved bootstrap
analysis that follows (Fig.~\ref{fig:modadd_bootstrap_loo}).

To quantify the statistical confidence of this crossing independently of the 
specific time-series trajectory, we performed bootstrap resampling and 
leave-one-scale-out (LOO) analysis on the phase-split $D$ values
(Fig.~\ref{fig:modadd_bootstrap_loo}).
Snapshots are split at the strict grokking boundary $t=0$: pre-grokking 
($t<0$) and post-grokking ($t>0$), then power-law fits are bootstrapped 
across seed--epoch samples.
The distributions are clearly resolved: $D_{\mathrm{pre}}=1.12\pm 0.02$, 
$D_{\mathrm{post}}=0.92\pm 0.02$, and the scale-free Gaussian synthetic baseline 
yields $D_{\mathrm{synth}}=1.00\pm 0.00$---the two empirical phases cross $D=1$ from opposite sides,
confirming that the crossing is statistically robust.
The LOO panel shows that removing any single intermediate scale shifts $D$ by
${<}0.02$, while excluding either extreme scale introduces a shift of at most
$0.08$; both phases remain well-separated under all six leave-one-out variants,
confirming the robustness of the phase separation.

\subsection{Negative controls and modulus robustness}
\label{sec:r3}
\begin{figure}[tp]
\centering
\includegraphics[width=\columnwidth]{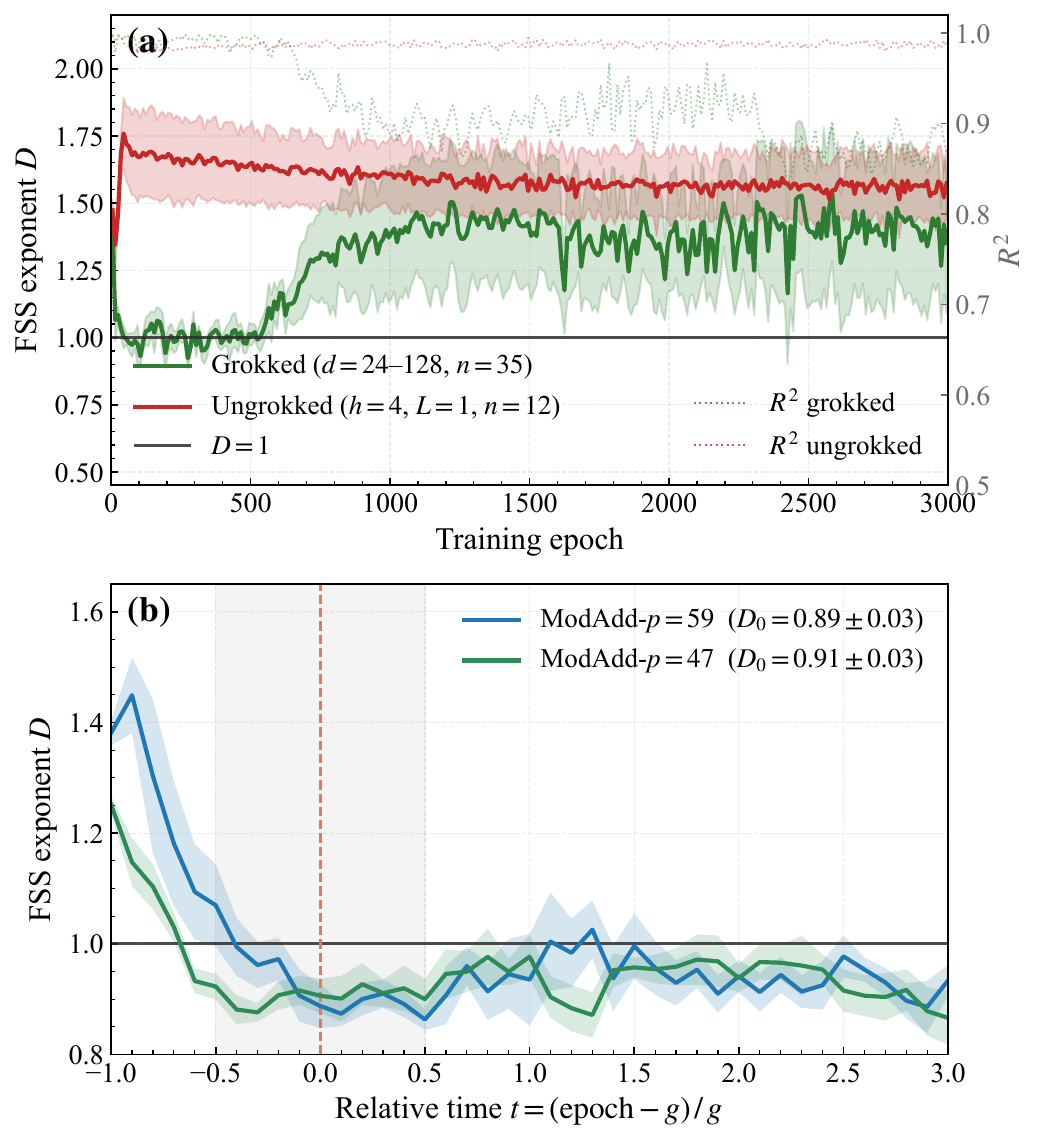}
\caption{\textbf{Grokking controls and modulus robustness in ModAdd.}
\textbf{(a)} Absolute-time FSS exponent $D(\mathrm{epoch})$ for grokked
vs.\ ungrokked ModAdd-59 runs ($h=4$, $L=1$, $p=59$).
\emph{Grokked:} the six canonical model widths $d=24$--$128$
($N\in\{6635,10363,20123,32955,67835,115003\}$), identical to the dataset
underlying all other FSS analyses in this paper.
\emph{Ungrokked:} all runs at the same architecture that failed
to generalize within the training budget---a broad negative control spanning
all available widths.
($n$: total training runs in each population.)
Dotted curves on the right axis show power-law fit quality $R^2$.
\textbf{(b)} Grokking-aligned FSS exponent $D(t)$ vs.\ normalized time
$t=(\mathrm{epoch}-g)/g$ for two modular primes.
ModAdd-$p=59$ (six canonical widths; same data as Fig.~\ref{fig:temporal_fss})
and the independent ModAdd-$p=47$ dataset (same canonical $h=4$, $L=1$
protocol) both cross $D=1$ at $t\approx 0$, with $D_0=0.89\pm 0.03$ and
$D_0=0.91\pm 0.03$, respectively, demonstrating robustness of the $D=1$ crossing across modular primes.
The shaded band marks the critical window $|t|\leq 0.5$.}
\label{fig:controls_universality}
\end{figure}

Panel~(a) of Fig.~\ref{fig:controls_universality} provides negative controls by
comparing ModAdd-59 trajectories that \emph{do} and \emph{do not} grok within
the training budget, both at the base architecture ($h=4$, $L=1$).
The \emph{grokked} set is restricted to the six canonical widths
$d=24$--$128$ ($N\in\{6635,10363,20123,32955,67835,115003\}$) that underpin all
FSS analyses in this paper; this ensures the data in panel~(a) is drawn from the
same models as the relative-time analysis of panel~(b) and
Figs.~\ref{fig:temporal_fss}--\ref{fig:modadd_bootstrap_loo}.
The \emph{ungrokked} set collects every run at the same architecture that failed
to generalize within the training budget, across all available widths, serving
as a broad negative control.
Ungrokked trajectories remain persistently supercritical ($D(t)>1$) and never
approach the post-transition regime reached by grokked runs.
Moreover, grokked and ungrokked trajectories diverge in $D$ well before
behavioral grokking (100--200~epochs in our runs), supporting the use of $D(t)$
as an early-warning diagnostic.

Panel~(a) also illustrates why grokking-aligned time is necessary: in absolute
time, the grokked population exhibits a non-monotonic $D(\mathrm{epoch})$
trajectory with no evident $D=1$ crossing.
This \emph{sequential phase-mixing} artifact arises because grokking epochs
span an order of magnitude across scales, pooling models in qualitatively
different dynamical phases at any fixed epoch.
The grokking-aligned bootstrap analysis of
Fig.~\ref{fig:modadd_bootstrap_loo} resolves this mixed signal into two
stationary phases: $D_{\mathrm{pre}}=1.12\pm 0.02$ (supercritical) and
$D_{\mathrm{post}}=0.92\pm 0.02$ (subcritical), with both separated from $D=1$
on opposite sides.

Panel~(b) of Fig.~\ref{fig:controls_universality} demonstrates that the $D=1$
crossing is robust to changing the modular prime.
Using the same canonical protocol ($h=4$, $L=1$, six widths $d=24$--$128$),
both ModAdd-$p=59$ and an independent ModAdd-$p=47$ dataset descend through
$D=1$ at $t\approx 0$, yielding $D_0=0.89\pm 0.03$ and $D_0=0.91\pm 0.03$,
respectively---demonstrating that the phase-transition signature is robust
across modular primes.

\subsection{Non-invasiveness controls}
\label{sec:r4}
\begin{figure}[tp]
\centering
\includegraphics[width=\columnwidth]{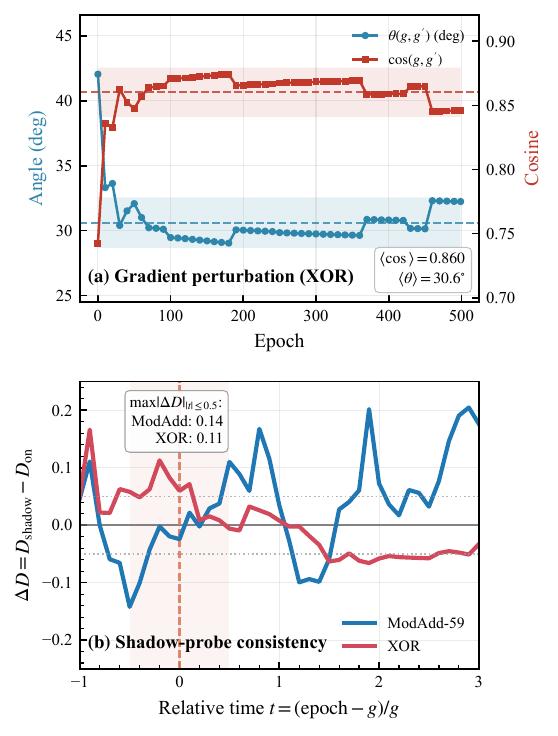}
\caption{\textbf{Probe impact and shadow-probe controls.}
\textbf{(a)} Under the Q90 setting, the TDU--OFC probe introduces appreciable
gradient rotations on XOR ($\langle\theta\rangle\simeq 30.6^{\circ}$,
$\langle\cos\rangle\simeq 0.86$).
\textbf{(b)} Shadow-probe consistency: $\Delta D = D_{\mathrm{shadow}} -
D_{\mathrm{on}}$ remains small in the critical window ($|t|\leq 0.5$), with
$\max|\Delta D| \leq 0.14$ (ModAdd) and $\leq 0.11$ (XOR).}
\label{fig:controls}
\end{figure}

Figure~\ref{fig:controls} tests whether the avalanche/FSS signatures could be
an artifact of measurement. Panel~(a) quantifies the local gradient rotation
introduced by the probe on XOR under shadow-probe conditions (Methods): the mean
angular deviation is $\langle\theta\rangle \simeq 30.6^{\circ}$
($\langle\cos\rangle \simeq 0.86$). Despite this, panel~(b) shows that the
macroscopic exponent is stable: a shadow-probe control
($\alpha_{\mathrm{train}}=0$) yields $\Delta D(t)=D_{\mathrm{shadow}}(t)-D_{\mathrm{on}}(t)$
with $\max|\Delta D|_{|t|\leq 0.5} \leq 0.14$ (ModAdd) and $0.11$ (XOR). The
invariance of $D$ under substantial gradient rotations supports the non-invasive
interpretation of the diagnostic.

\section{Discussion}

Our results support a physics picture of grokking as a \emph{dimensional 
phase transition} in gradient geometry. The key finding is not that $D \approx 1$ 
at grokking---uncorrelated Gaussian gradients also yield $D \approx 1$ on the same
topology---but that $D(t)$ \emph{dynamically crosses} this Gaussian/null baseline
from opposite directions in 
ModAdd (descending) and XOR (ascending). This opposite-direction attraction is consistent with \emph{attraction toward a shared
critical manifold at $D=1$}: the critical point acts as an attractor
in the space of optimization dynamics across the two systems studied,
independent of initial conditions set by task and architecture.

Distributional evidence complements the phase-resolved scaling: epoch-summed
avalanches are heavy-tailed with a characteristic cutoff $s_c$ that scales as
$s_c \sim N^{D_{\mathrm{cut}}}$ with $D_{\mathrm{cut}} \simeq 
1.02$ (Fig.~\ref{fig:cdf}).
Three independent lines of evidence exclude trivial origins of the 
$D \approx 1$ crossing:
(i) \emph{dynamical crossing} through the Gaussian/null baseline distinguishes 
grokking-localized criticality from static Gaussian baselines (which also 
yield $D \approx 1$);
(ii) \emph{shadow-probe controls} ($\alpha_{\mathrm{train}}=0$,
$\max|\Delta D| \leq 0.14$) confirm that $D(t)$ is robust under
appreciable gradient perturbations ($\sim 30^{\circ}$), ruling out
probe-induced artifacts (Fig.~\ref{fig:controls});
(iii) \emph{opposite crossing directions} across ModAdd (descending) and XOR 
(ascending) rule out task-specific coincidence and are consistent with a
shared attractor mechanism across the two systems studied.

\paragraph{Quasi-one-dimensional cascade geometry.}
The near-unity exponents across both systems---$D_0\approx 0.89$--$0.95$ 
near the transition and $D_{\mathrm{cut}}\simeq 1.02$ for distributional 
scaling---reflect a quasi-one-dimensional cascade geometry.
We note that $D_{\mathrm{cut}}\simeq 1.02$ is itself a time-average: 
because the analysis window spans the grokking transition, the aggregate 
CDF pools avalanches from both the pre-grokking regime ($D_{\mathrm{pre}}\simeq 1.12$) 
and the post-grokking regime ($D_{\mathrm{post}}\simeq 0.92$), 
bracketing the Gaussian baseline from opposite sides~\cite{wang2025grokking_xor}.

This is fundamentally different from the spatially extended two-dimensional 
avalanches in sandpile-type SOC models~\cite{bak1987self,pruessner2012self}, 
and is physically natural: gradient correlations during learning guide 
updates along low-dimensional solution manifolds in parameter 
space~\cite{saxe2014exact,jacot2018neural}, confining perturbation cascades 
to approximately one-dimensional paths through parameter space.
The same quasi-1D exponents ($D\approx 1.0$, $\gamma\approx 1.15$) were 
independently reported for XOR in Ref.~\cite{wang2025grokking_xor}, 
and the near-unity $D_{\mathrm{cut}}$ values here for ModAdd confirm 
this is not a task-specific coincidence---it is an intrinsic property 
of gradient correlation geometry in the studied systems.
Since $D$ reflects gradient field geometry rather than architectural details,
optimization-level interventions---such as gradient preprocessing or 
optimizer design---may be a fruitful avenue for further investigation.
By contrast, recent work on signal-propagation criticality reports distinct 
universality classes including directed percolation during general 
training~\cite{ghavasieh2025quasi,ghavasieh2025tunable}; our gradient-dynamics 
approach is complementary, specifically probing the transition geometry 
at the generalization event.

\paragraph{Toward a theoretical framework: critical manifold as attractor.}
Within a random-matrix--inspired picture, $D=1$ is the natural fixed point 
of isotropic, uncorrelated gradient diffusion~\cite{pruessner2012self}: 
when gradients lack structure, each parameter contributes independently and 
$s_{\max}\sim N^1$ follows from the linear superposition of uncorrelated 
degrees of freedom, as confirmed by our synthetic Gaussian 
controls~\cite{wang2025grokking_xor}.
Deviations from this baseline---whether upward (XOR) or downward 
(ModAdd)---therefore signal the emergence of \emph{structured, correlated} 
gradient dynamics, in the same sense that eigenvalue outliers departing from 
the random-matrix bulk signal low-rank coherent modes in spiked covariance 
models.
This systematic directional pattern---ModAdd descending, XOR ascending---suggests
that the $D=1$ critical manifold acts as an \emph{attractor} in the space of
gradient geometries.
Different task--architecture combinations appear to initialize the system at 
different positions in this space: ModAdd starts supercritical ($D > 1$), 
whereas XOR starts subcritical ($D < 1$), possibly reflecting differences 
in capacity-to-data ratios or gradient correlation structure.
Both are \emph{dynamically attracted} toward the same $D=1$ critical 
geometry at generalization.

This raises the question of \emph{why} the generalization transition
drives the system through $D=1$.
A useful interpretation emerges from a gradient-correlation flow picture.
In both systems studied, memorization and generalization correspond to
qualitatively distinct gradient correlation geometries that occupy
\emph{opposite sides} of the Gaussian diffusion baseline $D=1$:
for ModAdd, memorization is supercritical ($D_\mathrm{pre}>1$) while
generalization is subcritical ($D_\mathrm{post}<1$); for XOR the
ordering is reversed.
In this framework, $D=1$ represents the fixed point of isotropic,
uncorrelated gradient diffusion.
Structured gradient correlations move the system away from this neutral
baseline, placing memorization and generalization in distinct effective
correlation phases separated by the $D=1$ manifold.

Grokking therefore corresponds to a \emph{flow in gradient-correlation
space}, with $D=1$ acting as a separatrix analogous to a
renormalization-group fixed point that divides the two phases of
learning dynamics.
The crossing direction then reflects the initial gradient-correlation
regime set by task structure and architecture: systems initializing in
the supercritical regime (ModAdd) descend through $D=1$, while those
initializing in the subcritical regime (XOR) ascend.
The dimensional crossing $D=1$ therefore plays the role of a
\emph{transport critical point} in gradient space.
For $D>1$, gradient cascades are \emph{superdiffusive}: strongly
correlated updates propagate coherently across many parameters,
enabling extended exploration of solution space.
For $D<1$, cascades become \emph{subdiffusive}: updates are geometrically
confined to local parameter regions, reflecting constrained gradient-flow
structure.
$D=1$ corresponds to the Gaussian diffusion fixed point separating
superdiffusive and subdiffusive cascade regimes---the boundary at
which neither long-range correlation nor spatial confinement dominates.
Grokking corresponds to crossing this transport boundary, with the
direction (supercritical$\to$subcritical for ModAdd, reversed for XOR)
encoding the system's initial transport phase and connecting dimensional
criticality to the broader physics of anomalous diffusion in correlated
media~\cite{bak1987self,pruessner2012self}.

We therefore propose the \emph{Dimensional Grokking Principle}:
\emph{the grokking transition occurs when the effective gradient
cascade dimension $D(t)$ crosses the Gaussian diffusion baseline
$D=1$, approached from opposite directions by systems initializing
in distinct gradient-correlation phases.}
The crossing direction encodes the initial phase (supercritical or
subcritical), while the $D=1$ crossing itself is a consistent
signature of the generalization transition across the two systems studied.

This attractor hypothesis makes testable predictions: varying the 
capacity-to-data ratio should produce systems that cross $D=1$ from 
different directions or with different slopes, yet converge to similar 
$D$ values at the generalization transition.
This physics-based view complements circuit-level mechanistic 
interpretability~\cite{nanda2023progress} by providing a \emph{macroscopic, 
task-agnostic} diagnostic $D(t)$ that may extend to systems where 
circuit-level analysis is challenging.
Unlike behavioral metrics (loss, accuracy), $D(t)$ probes the geometry of 
optimization, enabling early-warning detection of generalization: 
grokked and ungrokked trajectories diverge in $D(t)$ at $\sim100$--$200$ 
epochs before behavioral grokking (e.g., $D \approx 1.02$ vs.\ $1.75$ 
at epoch 100, $R^2\geq 0.95$; separation $\Delta D \approx 0.73$), 
enabling threshold-based classification
of trajectory outcome well before the behavioral transition---a practically
useful property in settings where circuit-level interpretability is unavailable.
As emphasized above, the falsifiable signature is the temporally localized
crossing through the Gaussian/null baseline $D=1$---not static residence near
this value, which can arise trivially for unstructured gradients on the same
topology.

\paragraph{Falsifiable predictions.}
The critical-manifold attractor hypothesis makes concrete falsifiable 
predictions testable without circuit-level analysis.
First, tasks that initialize gradient geometry at $D>1$ should exhibit 
descending crossings (as in ModAdd), while those initializing at $D<1$ 
should ascend (as in XOR); intermediate initialization near $D=1$ 
would be expected to show a flatter, less decisive crossing.
Second, interventions that prevent grokking---including insufficient 
overparameterization or insufficient training budget (consistent with 
our ungrokked controls where runs remain at $D>1$ throughout training), 
and weight-decay suppression~\cite{power2022grokking}---should prevent 
the $D=1$ crossing entirely.
Independent structural confirmation comes from 
Ref.~\cite{wang2025grokking_xor}, where weight concentration (Gini 
coefficient of $|\theta|$) exhibits a transient $+25\%$ peak tightly 
synchronized with grokking---the same critical reorganization is thus 
visible from both the gradient-geometry ($D(t)$) and the weight-space 
perspective, providing mutually reinforcing evidence.
The precise universality class of the dimensional crossover identified 
here remains an open question warranting future investigation.

\paragraph{Limitations.}
The TDU--OFC cascade computation is $\mathcal{O}(N)$ per step, so scaling
to larger models will require approximate methods.
Our analysis covers two tasks (ModAdd and XOR) with two architectures
(Transformer and MLP); broader exploration across dataset complexities
and additional task families would further strengthen evidence for a
shared critical manifold.

\section{Conclusion}

We reported that the grokking transition---observed across Transformers
trained on modular addition and MLPs trained on XOR---corresponds to a
\emph{dimensional critical transition in gradient cascade geometry}: a
characteristic opposite-direction crossing of the Gaussian diffusion
baseline $D=1$ at generalization, with the crossing direction encoding
task structure and the critical manifold $D=1$ acting as a shared
attractor across the two systems studied.
Distributional avalanche statistics and shadow-probe controls provide
independent confirmation, establishing $D(t)$ as a macroscopic,
task-agnostic order parameter for the generalization transition.

Taken together, five convergent lines of evidence support dimensional
criticality at grokking: canonical behavioral delay, directional $D=1$
crossing, distributional finite-size scaling, negative controls on
ungrokked trajectories, and shadow-probe non-invasiveness.
Importantly, $D(t)$ separates grokked from ungrokked trajectories
roughly 100--200 epochs before behavioral grokking, providing a
practical early-warning signal of the impending generalization transition.
The temporally localized crossing through the Gaussian diffusion
baseline $D=1$---rather than static residence near this value---therefore
constitutes a falsifiable signature of grokking-specific criticality,
distinguishing the transition from both Gaussian baselines and
non-transitioning trajectories.

\section*{Data Availability}
The gradient snapshot data, analysis scripts, and model checkpoints
supporting the findings of this study are available from the corresponding
author upon reasonable request.

\begin{acknowledgments}
This work is supported by the National Key R\&D Program of China
(grant Nos. 2024YFA1611701, 2024YFA1611700).
\end{acknowledgments}

\bibliographystyle{apsrev4-2}
\bibliography{references.bib}

\end{document}